\documentclass[runningheads]{llncs}
\usepackage{graphicx}
\usepackage{amsmath,amssymb} 
\usepackage{array}
\newcolumntype{P}[1]{>{\centering\arraybackslash}p{#1}}
\newcolumntype{M}[1]{>{\centering\arraybackslash}m{#1}}
\newcolumntype{g}{>{\columncolor{Gray}}c}
\usepackage{color,colortbl}

\usepackage{newtxmath}
\usepackage{makecell}
\usepackage{multirow}
\usepackage{booktabs}
\usepackage{colortbl}
\usepackage{rotating}
\usepackage{hhline}
\usepackage{pifont}
\usepackage{bbm}
\usepackage{kotex}

\definecolor{Gray}{gray}{0.9}
\definecolor{olive}{RGB}{70, 105, 0}
\definecolor{brick}{RGB}{140, 0, 0}

\newcommand{\fref}[1]{Fig.~\ref{#1}}
\newcommand{\sref}[1]{Sec.~\ref{#1}}
\newcommand{\tref}[1]{Table~\ref{#1}}
\newcommand{\eref}[1]{Eq.~\ref{#1}}

\usepackage{xspace}
\makeatletter
\DeclareRobustCommand\onedot{\futurelet\@let@token\@onedot}
\def\@onedot{\ifx\@let@token.\else.\null\fi\xspace}

\def\eg{\emph{e.g}\onedot} 
\def\ie{\emph{i.e}\onedot}

\makeatother

\def\mF{\mathbf{F}}
\def\mA{\mathbf{A}}
\def\mAA{\tilde{\mathbf{A}}}
\def\mM{\mathbf{M}}
\def\mz{\mathbf{z}}
\def\loss{\mathit{\mathcal{L}}}
\def\fc{\text{fc}}
\def\expectation{\mathbb{E}}
\def\mx{\mathbf{x}}

\def\fg{\text{fg}}
\def\bg{\text{bg}}
\def\dfg{\text{dfg}}
\def\indicator{\mathbbm{1}}

\begin{document}
\pagestyle{headings}
\mainmatter

\def\ACCV20SubNumber{51}  

\title{In-sample Contrastive Learning and \\ Consistent Attention for Weakly Supervised Object Localization} 
\titlerunning{In-sample Contrastive Learning and Consistent Attention}

\author{Minsong Ki\inst{1} \and
Youngjung Uh\inst{2,3} \and
Wonyoung Lee\inst{3} \and Hyeran Byun\inst{1,3}}
\authorrunning{M. Author et al.}
\institute{Department of Computer Science, Yonsei University, Seoul, Republic of Korea \and
Department of Applied Information Engineering, Yonsei University, Seoul, Republic of Korea \and
Department of Artificial Intelligence, Yonsei University, Seoul, Republic of Korea
\email{\{kms2014,yj.uh,lwy8555,hrbyun\}@yonsei.ac.kr}}

\maketitle

\begin{abstract}
Weakly supervised object localization (WSOL) aims to localize the target object using only the image-level supervision. Recent methods encourage the model to activate feature maps over the entire object by dropping the most discriminative parts. However, they are likely to induce excessive extension to the backgrounds which leads to over-estimated localization. In this paper, we consider the background as an important cue that guides the feature activation to cover the sophisticated object region and propose contrastive attention loss. The loss promotes similarity between foreground and its dropped version, and, dissimilarity between the dropped version and background. Furthermore, we propose foreground consistency loss that penalizes earlier layers producing noisy attention regarding the later layer as a reference to provide them with a sense of backgroundness. It guides the early layers to activate on objects rather than locally distinctive backgrounds so that their attentions to be similar to the later layer. For better optimizing the above losses, we use the non-local attention blocks to replace channel-pooled attention leading to enhanced attention maps considering the spatial similarity. Last but not least, we propose to drop background regions in addition to the most discriminative region. Our method achieves state-of-the-art performance on CUB-200-2011 and ImageNet benchmark datasets regarding \texttt{top-1 localization accuracy} and \texttt{MaxBoxAccV2}, and we provide detailed analysis on our individual components. The code will be publicly available online for reproducibility.
\end{abstract}

\section{Introduction}

Fully supervised approaches have demonstrated excellent performance by training convolution neural network (CNN) with human annotations, \eg, bounding box for object localization, pixel-wise class labels for semantic segmentation \cite{chen2019towards,zhu2019feature,chen2019tensormask,robinson2020learning,li2018fast}. However, they cost huge human labor to obtain accurate annotations. Therefore, weakly supervised approaches that use only image-level supervision have received significant attention over the various computer vision tasks \cite{zhou2016learning,zhang2018adversarial,zhang2018self,choe2019attention,yang2020combinational,son2018forget,lee2020background}. Especially, weakly supervised object localization (WSOL) is a challenging task that pursues both classification and the localization of the target object where the training datasets provide only the class labels.

\begin{figure}[t]
\centering
\includegraphics[height=5.5cm, width=\columnwidth]{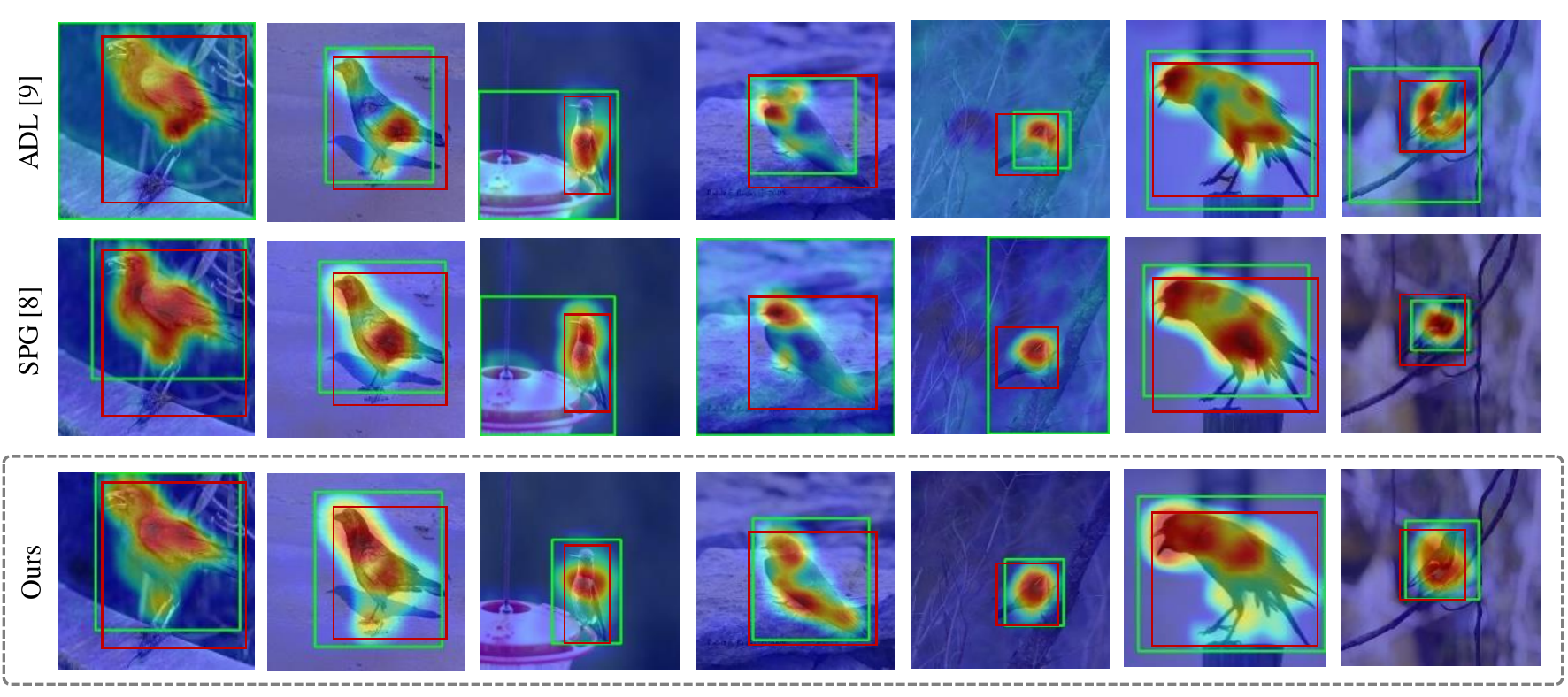} 
\caption{Comparison of methods for generating activation maps on the CUB \cite{wah2011caltech} dataset. We display the final results obtained by ADL \cite{choe2019attention} (first row), SPG \cite{zhang2018self} (second row), and our method (last row). The red boxes are the ground-truth and the green boxes are the predicted ones. Activation maps are illustrated in heatmap color scale.
ADL tries to activate more on less discriminative parts but ends in excessive extension to background. SPG tries to suppress background but still over-estimates the object regions. In contrast, our method covers the whole object delicately without extending to background.}
\label{fig:qual_comparison}
\end{figure}

For example, Zhou et al. \cite{zhou2016learning} generate class activation maps (CAM) using the classification model with a global average pooling (GAP). CAM highlights the class-specific discriminative regions in a given image~\cite{zhang2018adversarial,zhang2018self,choe2019attention,kumar2017hide}. The crucial pitfall of the activation maps is that it focuses on discriminative parts ({\it e.g.,} the head of a bird) rather than including the full extent of the object. 
To mitigate this limitation, recent methods \cite{choe2019attention,wei2017object,hou2018self} propose to erase the most discriminative parts by thresholding to spread out the activations to less discriminative regions. However, they are likely to induce excessive extension to the backgrounds which over-estimates bounding boxes (\fref{fig:qual_comparison}).

In this paper, we propose four ingredients for more accurate attention over the entire object: contrastive attention loss, foreground consistency loss, non-local attention block and dropped foreground mask. The contrastive attention loss draws the foreground feature and its erased version close together, and pushes the erased foreground feature away from the background feature (\sref{sect:caloss}).
It helps the learned representation to reflect only the object region rather than the backgrounds which are usually helpful for classification but harmful to localization. The foreground consistency loss penalizes disagreement of attentions between layers to provide early layers with a sense of backgroundness (\sref{sect:fcloss}). While usual low-level features are activated on locally distinctive regions (\eg, edges) regardless of the presence of the objects, adding foreground consistency loss boosts the activations on the object regions while suppressing the activations on the background regions. Furthermore, we apply the non-local attention blocks to produce enhanced attention maps considering the similarity between locations in a feature map (\sref{sect:attention}). It allows boosting weights on the regions having similar features with the most discriminative parts to pursue correct activation. Last but not least, we propose a dropped foreground mask which drops the background region as well as the most discriminative region. It prohibits the model from excessively spreading attention to backgrounds.

Our method achieves state-of-the-art performance in terms of the conventional \texttt{top-1 localization accuracy} and the \texttt{MaxBoxAccV2}~\cite{choe2020evaluating}. \\
\indent In summary, our main contributions are: 
\begin{itemize}
    \item[$\bullet$] We propose a contrastive attention loss that favors similarity between foreground feature and its dropped version and dissimilarity between the dropped foreground feature and background feature.
	\item[$\bullet$] We propose a foreground consistency loss that provides a sense of localization to earlier layers by guiding their features to be consistent with a high-level layer.
	\item[$\bullet$] We propose a dropped foreground mask which drops the background region and the most discriminative region.  
    \item[$\bullet$] Our method achieves state-of-the-art performance on CUB-200-2011 and ImageNet benchmark datasets in terms of \texttt{top-1 localization accuracy} and \texttt{MaxBoxAccV2}.
\end{itemize}

\section{Related Work}
\textbf{Weakly supervised object localization (WSOL).} Given only the class labels with the images, most of the WSOL methods train a classifier and extract CAM~\cite{zhou2016learning}. CAM indicates the strength of activation in every location in the feature map to stimulate the corresponding class~\cite{zhang2018adversarial,zhang2018self,choe2019attention,kumar2017hide}. Recent methods \cite{zhou2016learning,zhang2018adversarial,zhang2018self,choe2019attention,yang2020combinational,kumar2017hide} propose erasing the most discriminative region of the feature map to spread out the activations to the regions which are less discriminative but still in the object. Hide-and-Seek (HaS) \cite{kumar2017hide} divides a training image into a grid of evenly-divided patches and selects a random patch to be hidden. Adversarial complementary learning (ACoL) \cite{zhang2018adversarial} and attention-based dropout layer (ADL) \cite{choe2019attention} partially drop the most discriminative region by thresholding on the feature map. MEIL~\cite{mai2020erasing} runs two branches, one with erasing and one without erasing, and impose both branch with classification task. These approaches guide the models to discover previously neglected object regions. Our method steps further to consider background as a region to drop so that the model does not spread the activation excessively to the background.

Several methods have been proposed to suppress the background and localize the whole object. Zhang et al. \cite{zhang2018self} present self-produced guidance (SPG) that generates three pixel-wise masks (foreground, background, and undefined areas). Each mask is used as auxiliary supervision. However, it requires to find the optimal six hyperparameters for producing the three masks.
We also focus on the background but introduce simpler and more effective way. \\
\indent Yang et al. \cite{yang2020combinational} use a non-local block following every convolution-pooling block. While their non-local blocks are inserted within the main stem of the network, our non-local attention blocks are branch from the main stem and produce attention maps to be multiplied to the main convolutional features at chosen layers.

\noindent \textbf{Contrastive visual representation learning.}
Contrastive learning \cite{hadsell2006dimensionality} tries to distinguish similar and dissimilar pairs of samples by embedding the samples as feature representations. Recent self-supervised learning methods \cite{he2020momentum,chen2020simple} learn representations by maximizing agreement between differently augmented views of the same image. They also consider the different images to minimize the agreement for negative pairs. \\
\indent Inspired by \cite{he2020momentum,chen2020simple}, we define a contrastive prediction task for WSOL. 
Instead of building similar and dissimilar pairs of image samples, we regard the foreground region except for the most discriminative part (\ie, the dropped foreground) as an anchor, and build the positive pair with the original foreground and the negative pair with the background. Our contrastive objective does not require a large batch size or large queue because it finds the pairs within an image. Separating the foreground representation and the background representation is suitable for WSOL task.
\begin{figure}[!t]
\centering
\includegraphics[width=\columnwidth,height=5cm]{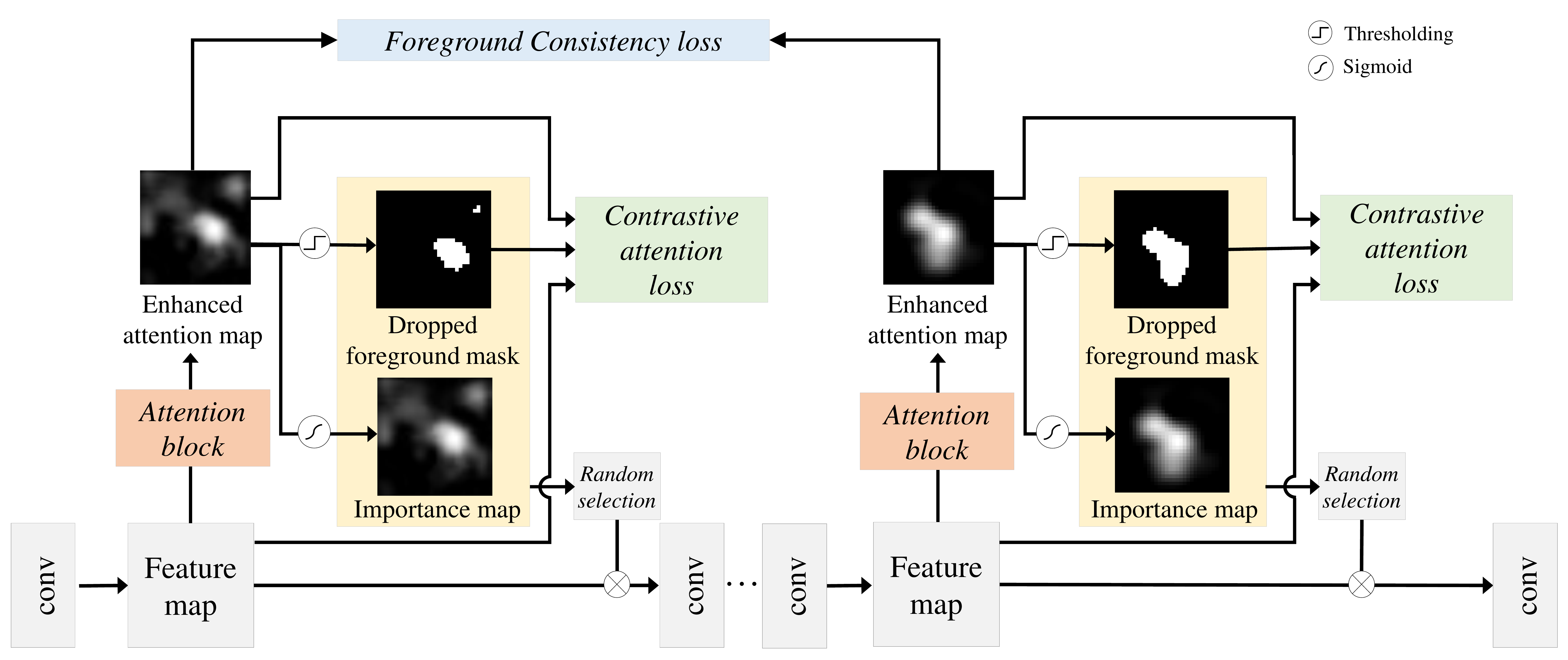} 
\caption{Overview of the proposed method. The non-local attention block generates the enhanced attention map reflecting the similarity between locations. We create a dropped foreground mask and an importance map using thresholding and sigmoid activation, respectively. The selected map is multiplied with the input feature to feed the next layer. Foreground consistency loss encourages the consistency between the early and last layer. We calculate the contrastive attention loss at each convolution layer where our non-local attention block is inserted.}
\label{fig:overview}
\end{figure}

\section{Proposed Method}
This section describes elements of the proposed method and how we employ them on the networks.
\subsection{Network overview}
As shown in \fref{fig:overview}, we augment classification network with the non-local attention blocks (\sref{sect:attention}) and train it with the contrastive attention loss (\sref{sect:caloss}) and the foreground consistency loss (\sref{sect:fcloss}). The non-local attention block receives a feature map $\mF$ and provides an enhanced attention map $\mA$ which becomes an importance map $\mAA$ through sigmoid activation and a dropped foreground mask $\mM_\dfg$ by thresholding (\eref{equ:dfg}). The dropped foreground mask or the importance map is randomly chosen based on a \texttt{drop\_rate} and the chosen one is applied to the input feature by pixel-wise multiplication (element-wise multiplication with broadcasting over the channel dimension). The importance map is not to be dropped but to be applied to the feature map. The dropped foreground mask encourages activation of the input feature on less discriminative parts except background to maximize classification accuracy without losing localization accuracy, while the importance map rewards higher activation on the most discriminative part. 
 
In the attention branch, the enhanced attention map and the dropped foreground mask from a non-local attention block are used to compute the contrastive attention loss. In addition, the enhanced attention maps from multiple non-local attention blocks are used to compute the foreground consistency loss.

\begin{figure}[t]
\centering
\includegraphics[width=8cm,height=2.8cm]{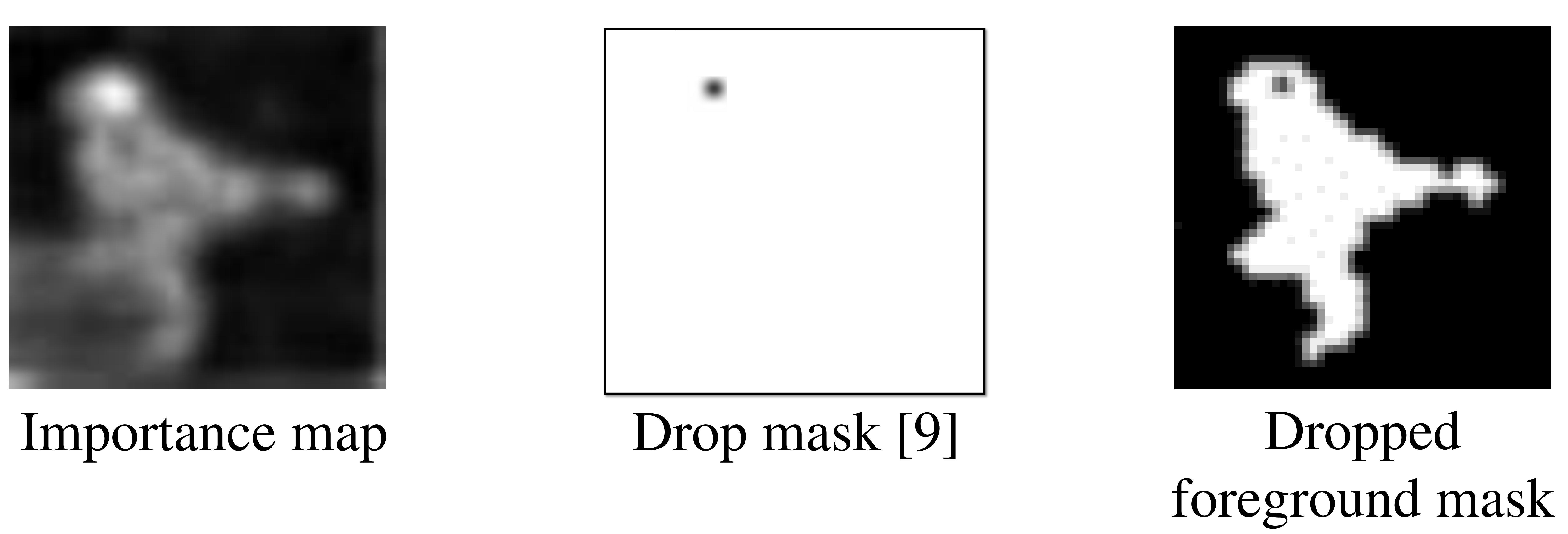} 
\caption{Examples of the importance map $\mAA$, the drop mask from \cite{choe2019attention}, and our dropped foreground mask $\mM_\dfg$.}
\label{fig:dfg}
\end{figure}

The differences with ADL \cite{choe2019attention} in the forward process are that we use the dropped foreground mask instead of drop mask and the attention map is produced by our non-local attention block instead of vanilla convolutional feature.
\fref{fig:dfg} illustrates the importance map, our dropped foreground mask, and the drop mask in \cite{choe2019attention}. Our dropped foreground mask $\mM_\dfg$ is defined by:
\begin{equation}
  \mM_\dfg =  \indicator[\mA < \theta_\fg] \land \indicator[\mA > \theta_\bg],
  \label{equ:dfg} 
\end{equation}
where $\indicator$ denotes a matrix with the same shape with the input having ones according to the logical operation, $\land$ denotes logical \texttt{and} operation, and $\theta$'s are the pre-defined thresholds.
Unlike the drop masks from ACoL~\cite{zhang2018adversarial} and ADL~\cite{choe2019attention}, our dropped foreground mask remedies excessive expansion of activation on the backgrounds by further erasing background regions in the mask. 

The contrastive attention loss and the foreground consistency loss are computed wherever the attention maps are extracted.

\begin{figure}[t]
\centering
\includegraphics[width=\columnwidth,height=6.9cm]{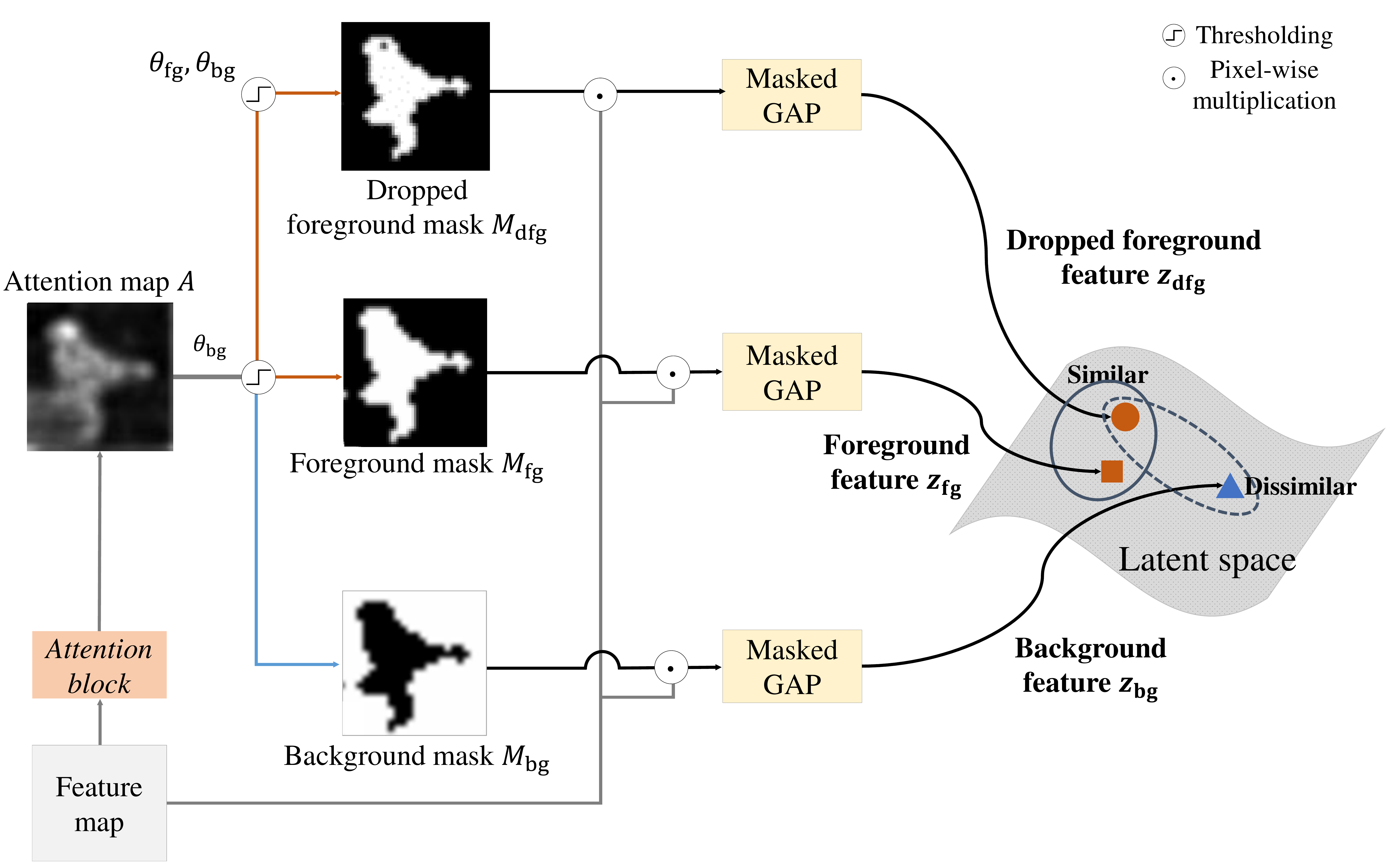} 
\caption{
The details of the contrastive attention loss, where $A$ denotes an enhanced attention map from a non-local attention block. We generate three maps and features to compare the similarity in embedding space. $\odot$ denotes pixel-wise multiplication. The contrastive attention loss is computed on an embedding space.}
\label{fig:caloss}
\end{figure} 

\subsection{Contrastive attention loss}\label{sect:caloss}
Contrastive loss \cite{he2020momentum} is a function whose value is low when a query is similar to its equivalent instance and dissimilar to its different instances. Likewise, we design a contrastive attention loss whose value is low when a dropped foreground feature $\mz_\dfg$ is similar to a foreground feature $\mz_\fg$ and dissimilar to a background feature $\mz_\bg$ (\fref{fig:caloss}). The features $\mz_\alpha$ are obtained by masked global average pooling of $\mF\odot\mM_\alpha$ where
\begin{equation}
\begin{split}
  \mM_\fg = \indicator[\mA > \theta_\bg], \\
  \mM_\bg = \indicator[\mA < \theta_\bg],
  \end{split}
  \label{equ:masks}
\end{equation}
and the masked global average pooling is spatial average pooling of the pixels whose value on the mask is $1$. Then, the contrastive attention loss is given by
\begin{equation}
  \loss_{\text{ca}} = \expectation_{\mx}[[(d(z_{\text{dfg}}, z_{\text{fg}}) - d(z_{\text{dfg}}, z_{\text{\bg}}) + m]_+],
  \label{equ:caloss}
\end{equation}
where $[\cdot]_+=\text{max}(\cdot, 0)$ and $d(\cdot, \cdot)$ denotes $L_2$ distance in auxiliary 128-dimensional embedding by 1x1 convolution. $m$ denotes the margin.

Our contrastive attention loss guides the attention map to spread until it reaches boundary because including backgrounds in the attention map is penalized by the dissimilarity term. In addition, the similarity term favors homogenous features between the most discriminative part and less discriminative parts in the foreground region.
Our contrastive attention loss does not require mining positive and negative samples as in triplet loss \cite{chen2017beyond} nor managing large negative samples \cite{he2020momentum,chen2020simple}. Since we regard the masked features $\mz_{\dfg}, \mz_{\fg}$ and $\mz_{\text{bg}}$ from one image as an anchor, a positive sample and a negative sample, respectively.

\subsection{Foreground consistency loss}\label{sect:fcloss}
Attention maps roughly are the magnitude of activation on every location. Convolutions in early layers activate more on locally distinctive regions such as edges and corners \cite{zeiler2014visualizing}, without inspecting the entire extent of objects due to their limited receptive field. To relieve this problem, we propose a foreground consistency loss that encourages attention maps from early layers to resemble later layers (\fref{fig:overview}).

Let ${A}_{i}$ and ${A}_{j}$ are the attention maps from early and later layers, respectively. Then we define the foreground consistency loss as:

\begin{equation}
  \loss_{fc} = {\begin{Vmatrix}
{A}_{i} - {A}_{j}
\end{Vmatrix}}^2_2,
  \label{equ:fcloss}
\end{equation}
where $\|\cdot\|_2$ denotes $L_2$ norm of a matrix.

Gradients from the foreground consistency loss only run through the early layer to achieve the abovementioned goal. It reduces the noisy activations outside the object and boosts activations in the object.

\subsection{Non-local attention block}\label{sect:attention}
In order to provide additional capacity for the network to produce a correct attention map, we employ non-local block \cite{wang2018non} instead of average channel pooling of the convolutional features~\cite{choe2019attention,baek2020psynet}. Given a feature map, our non-local attention block embeds it into three different embeddings and outputs spatial summation of the third one weighted by similarity between the first two embeddings. Then the enhanced attention map is defined by its channel-pooled result.

Specifically, the block receives a feature map $x\in {R}^{C\times H\times W}$ from a convolution layer.
For simplicity, we omit the mini-batch dimension. We define $f(x), g(x)\in {R}^{\tilde{C}\times H\times W}, z(x)\in {R}^{C\times H\times W}$ that use 1 x 1 convolution layer for embedding. Then, $f(x), g(x)$ and $z(x)$ are reshaped to $f(x), g(x)\in {R}^{\tilde{C}\times HW}, z(x)\in {R}^{C\times HW}$, respectively. 

The enhanced attention map $A$ is given by:

\begin{equation}
  {A} = \expectation_C[\text{Softmax}({f(x)}^{T}g(x)) \odot z(x)],
  \label{equ:nonloc}
\end{equation}
where $\expectation_C$ denotes average pooling over the channel dimension.

The non-local attention block produces the enhanced attention map regarding similarities between locations. It unleashes the receptive field of the layer and provides an additional clue for deciding where to attend.
Our non-local attention block is different from \cite{yang2020combinational} in that we organize it only when generating several enhanced attention maps. Yang et al.~\cite{yang2020combinational} apply the non-local module to all layers in the main branch with residual connection. 

\subsection{Training and Inference}\label{sect:losses}
We train the base network and non-local attention block with the full objective:

\begin{equation}
  {\mathcal{L}}_{total} = {\mathcal{L}}_{cls} + {\mathcal{L}}_{ca} + {\mathcal{L}}_{fc}
  \label{equ:totalloss}
\end{equation}
\smallskip

We employ a GAP layer at the end of the network to produce softmax output $\hat{y}$ and compute classification loss given the one-hot ground truth label $y$:
 
\begin{equation}
  {\mathcal{L}}_{cls} = \text{CrossEntropy}(\hat{y}, y)
  \label{equ:clsloss}
\end{equation}

All network weights are updated wherever all losses send their gradients towards the input, while the foreground consistency loss does not convey gradients to its reference layer.

Our non-local attention block is applied only during training and deactivated in the testing phase. The input image goes through only the vanilla model to produce the class assignment. 
Then we follow \cite{choe2020evaluating} to extract the heatmap which leads to the bounding boxes by thresholding and its connected multi-contour.

\section{Experiments}\label{sect:experiments}
\subsection{Experimental setup}\label{sect:exsetup}

\textbf{Datasets.} We evaluate the proposed method on two benchmark datasets: CUB-200-2011 \cite{wah2011caltech} and ILSVRC \cite{russakovsky2015imagenet} (ImageNet) for WSOL task, from which only the image-level labels are used in training. Many weak-supervision methods have used full supervision to some extent, directly or indirectly for hyperparameter tuning. Since the amount of full supervision used for hyperparameter tuning is not consistent, it has been ambiguous using the previous evaluation metric for a fair comparison. We follow the recent evaluation metric \cite{choe2020evaluating} which fixes the amount of full supervision only for  hyperparameter search. 
Each dataset is divided into three subsets: 
\texttt{train-weaksup}, \texttt{train-fullsup} and \texttt{test}. The \texttt{train-weaksup} includes images only with the class labels for training. The \texttt{train-fullsup} contains images with full supervision, which has bounding boxes as well. It is left free for the users to use the \texttt{train-fullsup} for hyperparameter search. They collected five images per class (total 1,000 images) from Flickr for CUB experiments, and ten images per class (total 10,000 images) from ImageNetV2 \cite{recht2019imagenet} for ImageNet experiments, respectively. The \texttt{test} split for the final number is the same as the standard WSOL settings on CUB and ImageNet experiments~\cite{zhang2018adversarial,zhang2018self,choe2019attention,choe2020evaluating,yun2019cutmix}. In the CUB dataset, there are 5,994 images for training and 5,794 for testing from 200 bird species. ImageNet consists of 1.2M training images and 10K test images for 1,000 classes. All experimental analyses of the proposed method are conducted on the \texttt{test} split of the two abovementioned datasets. 

\smallskip
\noindent\textbf{Evaluation metrics.}
We use \texttt{top-1 classification accuracy} and  \texttt{top-1 localization accuracy}, and \texttt{MaxBoxAccV2} \cite{choe2020evaluating}. 

\texttt{Top-1 classification accuracy} is the ratio of correct classification. The conventional \texttt{top-1 localization accuracy} measures ratio of the samples with the right class and the bounding box of IoU greater than 0.5. 

\texttt{MaxBoxAcc} measures ratio of the samples with the correct box, while the correctness is defined by an IoU criterion $\delta$ at the optimal activation threshold. \texttt{MaxBoxAccV2} averages \texttt{MaxBoxAcc} at three IoU criterions $\delta \in \{30, 50, 70\}$ to address diverse demands for localization fineness. It is similar to the common GT-known metric but differs in that it evaluates on \textit{three IoUs} by extracting the bounding box with \textit{the optimal score map threshold}. We use \% symbol as a \textit{percent point} for mentioning differences on comparisons.

\smallskip
\noindent\textbf{Implementation details.}
We build the proposed method upon three CNN backbones: VGG16 \cite{simonyan2014very}, InceptionV3 \cite{szegedy2016rethinking}, and Resnet50 \cite{he2016deep}. 
We need three hyperparameters: \texttt{drop\_rate} for randomly choosing the importance map or the dropped foreground mask, $\theta_\fg$ and $\theta_\bg$ for thresholding.
The threshold $\theta_\fg$ is set to the maximum intensity of $\mA$ times pre-defined ratio $\gamma_\fg$. The $\theta_\bg$ is set to average intensity of $\mA$ times pre-defined ratio $\gamma_\bg$. 
The specific values of the hyperparameters for each backbone are shown in \tref{table:hyperparam}.

The layers, from which the enhanced attention maps are extracted, are chosen to be the same with the baseline method \cite{choe2019attention}. We also calculate our contrastive attention loss and foreground consistency loss for all layers where attention maps are produced. We set the batch size to 32, weight decay to 0.0001, margin $m$ to 1. The initial learning rate and the momentum of the SGD optimizer are set to 0.001 and 0.9, respectively. We start from loading weights from the model pre-trained on the ImageNet classification \cite{russakovsky2015imagenet} and then fine-tuned the network. Our model is implemented using PyTorch and trained using two NVIDIA GeForce RTX 2080 Ti GPUs for approximately three hours. The input images are randomly cropped to 224 $\times$ 224 pixels after being resized to 256 $\times$ 256 pixels. During testing phase, we directly resize the input images to 224 $\times$ 224. 

\begin{table}[t]
\begin{center}
\caption{Hyperparameters (\texttt{drop\_rate}, $\gamma_\fg$, $\gamma_\bg$) for each backbone.}
\label{table:hyperparam}
\begin{tabular}{M{2.5cm}|M{1.7cm}M{1.7cm}M{1.7cm}}
\toprule
Backbone  & \texttt{drop\_rate} &  $\gamma_\fg$ & $\gamma_\bg$  \\
\noalign{\smallskip}\hline
VGG \cite{simonyan2014very} & 0.33  & 0.72   & 1.2  \\
InceptionV3 \cite{szegedy2016rethinking} & 0.69  & 0.86   & 1.2 \\
ResNet50 \cite{he2016deep} & 0.85  & 0.95  & 1.2  \\
\bottomrule
\end{tabular}
\end{center}
\end{table} 

\subsection{Ablation study}\label{sect:analysis}
We first show detailed experiments to validate effectiveness of each component. We fix the ResNet50 \cite{he2016deep} as a backbone and add or remove each component. The experiments are performed on the CUB \texttt{test} split. 
The difference in performance in \% represents percent points.  

\smallskip
\noindent \textbf{Ablation of the proposed losses.}
\tref{table:ablation} shows that both the contrastive attention loss and the foreground consistency loss are the crucial element for the improved performance. Ours without the contrastive attention loss achieves 2\% lower performance than the full setting. The loss has positive effect on all three IoU thresholds.
The foreground consistency loss also plays an important role of improvement by 0.79\%. It especially boosts the accuracy at IoU 0.7. We suggest that the loss helps precisely estimating the location of the object in the early layer by providing the hints from the later layer. 
Using the both losses leads to balanced improvements over all IoU thresholds.
In addition, contrastive attention loss with the normalized temperature-scaled cross-entropy loss \textit{(NT-Xent)} \cite{he2020momentum,chen2020simple} also shows improvements to some extent. Its result can be found in the supplementary material.
\smallskip
\begin{table}[b]
\begin{center}
\caption{The ablation study for each element of our method on Resnet50 \cite{he2016deep} backbone in terms of \texttt{MaxBoxAccV2}. Contrastive: contrastive attention loss. $\loss_\fc$: foreground consistency loss. Non-local: non-local attention block. $\mM_\dfg$: dropped foreground mask. All elements contribute to the performance improvement.}
\label{table:ablation}
\begin{tabular}{l|M{1.7cm}M{1.7cm}M{1.7cm}g}
\toprule
\multirow{2}*{Methods}  & 
\multicolumn{4}{c}{\texttt{MaxBoxAccV2}@IoU (\%)} \\ & 0.3  & 0.5   & 0.7  & Mean \\
\hline
Baseline \cite{choe2019attention} & 91.82 & 64.78 & 18.43 & 58.34 (-4.86)  \\
Ours w/o contrastive  & 94.79 & 70.84 & 17.98 & 61.20 (-2.00)\\
Ours w/o $\loss_\fc$ & \textbf{96.51} & 72.14 & 18.57 & 62.41 (-0.79) \\
Ours w/o non-local & 95.80 & 71.62 & 20.32 & 62.58 (-0.62) \\
Ours w/o $\mM_\dfg$ & 96.42 & 72.21 & 17.62 & 62.08 (-1.12) \\
Ours (full) &  96.18 & \textbf{72.79}  & \textbf{20.64} & \textbf{63.20} \\
\bottomrule
\end{tabular}
\end{center}
\end{table}

\noindent \textbf{Effectiveness of the non-local attention block.} If we use the vanilla attention map which is the channel-pooled result of the convolutional feature, the performance drops by 0.62\% (the fourth row in \tref{table:ablation}). It shows that considering the relationship between pixels in feature map helps localizing where to attend.
\smallskip

\noindent \textbf{Effectiveness of the dropped foreground mask.} Here we validate the effectiveness of replacing the drop mask \cite{choe2019attention} with the dropped foreground mask $\mM_\dfg$. Without the replacement, the model achieves 1.12\% lower performance than the full setting (the fifth row in \tref{table:ablation}). Also, only replacing the drop mask with the dropped foreground mask improves the performance of the baseline \cite{choe2019attention} by 0.87\%.
We suppose that the dropped foreground mask improves ours more than the baseline because the additional two losses and the non-local attention block provide an extra guide for better importance map.
\\

\noindent \textbf{Location of our attention block.} We investigate the influence of where to insert our non-local attention block on VGG16 \cite{simonyan2014very}, and report the results in \tref{table:where}. 
The \texttt{conv\_5\_3} layer is fixed as the reference layer for the foreground consistency and its preceding layers are added one by one cumulatively. The setting with top three layers, which is the same as the baseline \cite{choe2019attention}, achieves the best performance in terms of \texttt{MaxBoxAccV2}. 
Adding the attention blocks on \texttt{pool\_1} and \texttt{pool\_2} layers decreases the performance. We suppose that the reason is their small receptive field which leads to noisy activations on extremely locally salient regions. Hence, we do not use the attention mechanism on the two earliest layers.
\begin{table}[t]
\begin{center}
\caption{Performance comparison regarding at which layer to insert our attention block. The contrastive attention loss and the foreground consistency loss are in use for all cases. The \texttt{conv\_5\_3} layer is fixed as the reference layer and its performance is left empty because the foreground consistency loss requires at least two layers. We add the layers from later to earlier and report their performance in a cumulative setting.}

\label{table:where}
\begin{tabular}{l|c|c} 
\toprule
Location   & ~~\texttt{Top-1 classification}~~ &  ~~\texttt{MaxBoxAccV2}~~ \\
\hline
\texttt{conv\_5\_3}  & - & - \\
~~~$\mathit{+}$ \texttt{pool\_4} & {74.23} & 64.67 \\
~~~$\mathit{+}$ \texttt{pool\_3} &  73.35 & \textbf{66.72} \\
~~~$\mathit{+}$ \texttt{pool\_2} & 66.26 & 63.25 \\
~~~$\mathit{+}$ \texttt{pool\_1} & 62.68 & 62.18 \\
\bottomrule
\end{tabular}
\end{center}
\end{table}
\begin{table}[!b]
\begin{center}
\caption{\texttt{MaxBoxAccV2} \cite{choe2020evaluating} comparison with the state-of-the-art methods. The results for each backbone represent the average of the three IoU thresholds 0.3, 0.5, and 0.7. VGG: VGG16 \cite{simonyan2014very}. Inc: InceptionV3 \cite{szegedy2016rethinking}. Res: ResNet50 \cite{he2016deep}. The best and the second best entries in a column are marked in boldface and italic, respectively.}
\label{table:maxboxaccv2}
\begin{tabular}{l|M{1.2cm}M{1.2cm}M{1.2cm}g|M{1.2cm}M{1.2cm}M{1.2cm}g}
\toprule
\multirow{2}*{Methods}  & \multicolumn{4}{c|}{ImageNet}  &  \multicolumn{4}{c}{CUB-200-2011} \\ 
 & VGG & Inc & Res & Mean & VGG & Inc & Res & Mean \\
\hline
CAM \cite{zhou2016learning}& 60.0 & 63.4 & 63.7&  62.4 &  63.7 & 56.7 & 63.0 & 61.1 \\ 
HaS \cite{kumar2017hide}& 60.6 & 63.7 & 63.4 &  \textit{62.6} &  63.7 & 53.4 & 64.7 & 60.6 \\ 
ACoL \cite{zhang2018adversarial}& 57.4 &  63.7 & 62.3 &  61.2 &  57.4 & 56.2 & \textbf{66.5} & 60.0 \\   
SPG \cite{zhang2018self}& 59.9 & 63.3 & 63.3 &  62.2 &  56.3 & 55.9 & 60.4 & 57.5 \\ 
ADL \cite{choe2019attention}& 59.8 &  61.4 & 63.7 & 61.7 &  66.3 & 58.8 & 58.4 & \textit{61.1} \\ 
CutMix \cite{yun2019cutmix}& 59.4 &  \textbf{63.9} & 63.3 & 62.2 & 62.3 & 57.5 & 62.8 & 60.8 \\ 
\hline\noalign{\smallskip}
Ours & \textbf{61.3} & 62.8 & \textbf{65.1} & \textbf{63.1} &  \textbf{66.7} & \textbf{60.3} & 63.2 & \textbf{63.4} \\ 
\bottomrule
\end{tabular}
\end{center}
\end{table}
\subsection{Comparison with state-of-the-art methods}
We compare our method with the state-of-the-art WSOL methods in terms of the \texttt{MaxBoxAccV2} \cite{choe2020evaluating}, \texttt{top-1 localization} and \texttt{top-1 classification accuracy}. 
\smallskip

\noindent \textbf{MaxBoxAccV2 \cite{choe2020evaluating}.} 
\tref{table:maxboxaccv2} shows comparison of \texttt{MaxBoxAccV2} across all competitors on ImageNet and CUB. Our method outperforms all existing methods in terms of \texttt{MaxBoxAccV2} (Mean) and most of backbone choices. \tref{table:maxboxaccv2_detail} shows detailed comparison with the runner-up methods of each dataset. Our method boosts performance especially when IoU criterions are 0.5 and 0.7 except when Inception network is the backbone. Our method exhibits the largest improvement when employed on ResNet backbone. 
\smallskip

\begin{table}[!t]
\begin{center}
\caption{Detailed \texttt{MaxBoxAccV2} \cite{choe2020evaluating} comparison with the runner-up methods on each dataset. 
We compare ours and the second best methods on each dataset and each backbone in terms of \texttt{MaxBoxAccV2} including individual measures on the three IoU criterions.
Mean indicates that the average value of the three IoU thresholds. VGG: VGG16 \cite{simonyan2014very}. Inc: InceptionV3 \cite{szegedy2016rethinking}. Res: ResNet50 \cite{he2016deep}. Bold texts denote the best performance in each column.}
\label{table:maxboxaccv2_detail}
\begin{tabular}{c|M{1.1cm}M{1.4cm}|M{1cm}|M{1.9cm}M{1.9cm}M{1.9cm}g}
\toprule
 & \multirow{2}*{Method} & \multirow{2}*{Backbone} & \multirow{2}*{Top-1} &\multicolumn{4}{c}{\texttt{MaxBoxAccV2}@IoU (\%)} \\ & & & & 0.3 & 0.5 & 0.7 & Mean \\ 
\hline 
\multirow{6}*{\rotatebox[origin=c]{90}{CUB-200-2011}} & ADL & VGG & 54.95 & 97.72 & 78.06 & 23.04 & 66.28 \\
& Ours & VGG & 73.35 & 96.20 (\textcolor{red}{$-$1.5}) & 77.20 (\textcolor{red}{$-$0.8}) & 26.75 (\textcolor{blue}{$+$3.7}) & \textbf{66.72} (\textcolor{blue}{$+$0.5}) \\
\cline{2-8}
& ADL & Inc & 41.03 & 93.77 & 65.79 & 16.86 & 58.81 \\
& Ours & Inc  & 64.01 & 95.89 (\textcolor{blue}{$+$2.1}) & 67.93 (\textcolor{blue}{$+$2.2}) & 17.20 (\textcolor{blue}{$+$0.4}) & \textbf{60.34} (\textcolor{blue}{$+$1.2}) \\ 
\cline{2-8}
& ADL & Res & 66.60 & 91.82 & 64.76 & 18.43 & 58.34 \\  
& Ours & Res  & 80.35 & 96.18 (\textcolor{blue}{$+$4.3}) & 72.79 (\textcolor{blue}{$+$8.0}) & 20.64 (\textcolor{blue}{$+$2.2}) & \textbf{63.20} (\textcolor{blue}{$+$4.9}) \\ 
\hline\noalign{\smallskip}
\hline\noalign{\smallskip}
\multirow{6}*{\rotatebox[origin=c]{90}{ImageNet}} & HaS & VGG & 68.26 & 80.72 & 62.15 & 38.89 & 60.59 \\
& Ours & VGG &  69.21 & 81.45 (\textcolor{blue}{$+$0.7}) & 63.20 (\textcolor{blue}{$+$1.1}) & 39.35 (\textcolor{blue}{$+$2.2}) & \textbf{61.33} (\textcolor{blue}{$+$0.8}) \\ 
\cline{2-8}
& HaS & Inc & 69.07 & 83.95 & 66.27 & 40.94 & \textbf{63.72} \\
& Ours & Inc  & 71.31 & 82.44 (\textcolor{red}{$-$1.5}) & 65.21 (\textcolor{red}{$-$1.0}) & 40.87 (\textcolor{red}{$-$0.1}) & 62.84 (\textcolor{red}{$-$0.9}) \\ 
\cline{2-8}
& HaS &  Res & 75.39 & 83.71 & 65.22 & 41.26 & 63.40 \\
& Ours &  Res  & 76.54 & 84.26 (\textcolor{blue}{$+$0.5}) & 67.62 (\textcolor{blue}{$+$2.4}) & 43.58 (\textcolor{blue}{$+$2.3}) & \textbf{65.15} (\textcolor{blue}{$+$1.7}) \\
\bottomrule
\end{tabular}
\end{center}
\end{table}
\setlength{\tabcolsep}{1.4pt}


\smallskip

\noindent \textbf{Top-1 localization accuracy.} \texttt{Top-1 localization accuracy} on the ImageNet and CUB datasets is shown in \tref{table:top1loc}. Our model outperforms the state-of-the-art methods on most settings. Note that we do not perform hyperparameter tuning using the \texttt{train-fullsup} split following the competitors for a fair comparison. 
\begin{table}[!t]
\begin{center}
\caption{Conventional \texttt{Top-1 localization accuracy} comparison with the state-of-the-art methods. 
The values are taken from their respective papers. Bold texts denote the best performance in each backbone network. 
}
\label{table:top1loc}
\begin{tabular}{l|M{1.6cm}M{1.6cm}M{1.6cm}|M{1.6cm}M{1.6cm}M{1.6cm}}
\noalign{\smallskip}\toprule
\multirow{2}*{Methods}  & \multicolumn{3}{c|}{ImageNet}  &  \multicolumn{3}{c}{CUB-200-2011} \\  
\cline{2-7}
& VGG & Inc & Res & VGG & Inc & Res \\
\hline
CAM \cite{zhou2016learning}& 42.8 & 46.3 & - &  37.1 &  43.7 & 49.4  \\ 
HaS \cite{kumar2017hide}& - & - & - &  - &  - & -  \\ 
ACoL \cite{zhang2018adversarial}& 45.8 & - & - &  45.9 &  - & - \\   
SPG \cite{zhang2018self}& - & 48.6 & - &  - &  46.6 & - \\ 
ADL \cite{choe2019attention}& 44.9 & 48.7 & - &  52.4 &  53.0 & - \\ 
CutMix \cite{yun2019cutmix}& 43.5 & - & 47.3 &  - &  52.5 & 54.8 \\ 
MEIL \cite{mai2020erasing}& 46.8 & \textbf{49.5} & - &  57.5 &  - & - \\
\hline\noalign{\smallskip}
Ours &  \textbf{47.2} & 49.3 & \textbf{48.4} &  \textbf{57.5} &  \textbf{56.1} & \textbf{56.1} \\ 
\bottomrule
\end{tabular}
\end{center}
\end{table}

\smallskip

\noindent \textbf{Top-1 classification accuracy.}
\tref{table:top1cls} compares our method with the state-of-the-art methods in terms of \texttt{top-1 classification accuracy}. While some other methods compromise classification accuracy for improving locazliation, our method achieves the best \texttt{MaxBoxAccV2} and localization accuracy without damaging the classification accuracy.
\begin{table}[!t]
\begin{center}
\caption{\texttt{Top-1 classification} performance of the state-of-the-art methods. Hyperparameters for each method are optimally selected for the localization performances on \texttt{train-fullsup} split. 
Bold texts denote the best performance. MEIL does not provide code for reproduction and its values are taken from the paper. Other values are reproduction from \cite{choe2020evaluating}.}
\label{table:top1cls}
\begin{tabular}{l|M{1.2cm}M{1.2cm}M{1.2cm}g|M{1.2cm}M{1.2cm}M{1.2cm}g}
\toprule
\multirow{2}*{Methods}  & \multicolumn{4}{c|}{ImageNet}  &  \multicolumn{4}{c}{CUB-200-2011} \\ 
 & VGG & Inc & Res & Mean & VGG & Inc & Res & Mean \\
\hline
CAM \cite{zhou2016learning}& 66.48 & 70.56 & 75.05 & 70.70 & 50.10 & 70.70 & 71.50 & 64.10 \\ 
HaS \cite{kumar2017hide}& 68.26 & 69.07 & 75.39 & 70.91 & 75.90 & 64.50 & 69.70 & 70.10  \\ 
ACoL \cite{zhang2018adversarial}& 64.55 & 71.81 & 73.09 & 69.82 & 71.80 & 71.50 & 71.10 & 71.40 \\ 
SPG \cite{zhang2018self}& 67.76 & 71.12 & 73.26 & 70.71 & 72.10 & 46,20 & 50.50 & 56.30 \\ 
ADL \cite{choe2019attention}& 67.58 &  61.17 & 71.99 & 66.91 & 55.00 & 41.00 & 66.60 & 54.20 \\ 
CutMix \cite{yun2019cutmix}& 66.36 & 69.16 & 75.71 & 70.41 & 48.40 & 71.00 & 73.00 & 64.10 \\ 
MEIL \cite{mai2020erasing}& 70.27 & 73.31 & - &- & 74.77 & 74.55 & - & \\ 
\hline\noalign{\smallskip}
Ours & 69.21 & 76.54 & 71.31 & \textbf{72.35} & 73.40 & 64.00 & 80.40 & \textbf{72.60} \\
\bottomrule
\end{tabular}
\end{center}
\end{table}
\subsection{Qualitative results}
\begin{figure}[!t]
\centering
\includegraphics[width=\columnwidth,height=8.4cm]{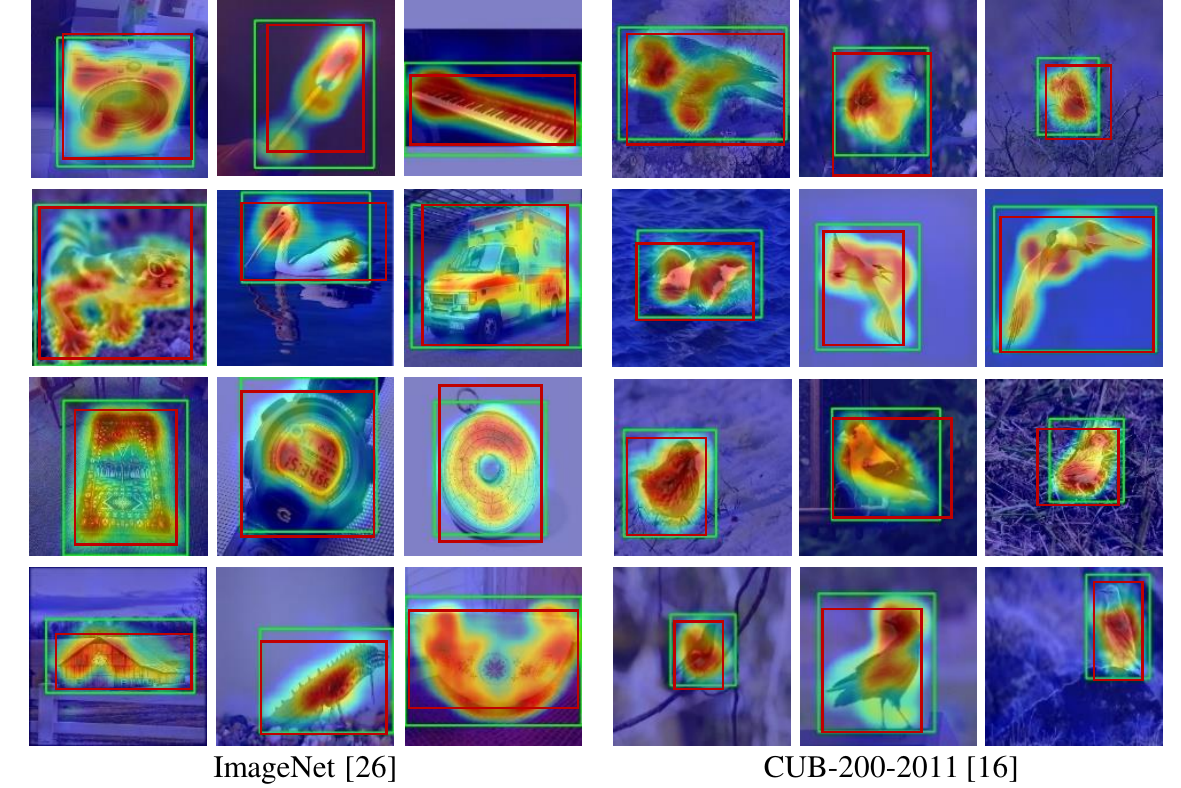} 
\caption{Qualitative examples of activation map and localization produced by our model on the ImageNet and CUB \texttt{test} split. The red boxes are the ground-truth and the green boxes are the predicted ones. These maps output with colors ranging from red (higher importance) to blue (lower importance like a background).}
\label{fig:figure4}
\end{figure}

\fref{fig:qual_comparison} compares activation maps and estimated bounding boxes from ADL~\cite{choe2019attention}, SPG~\cite{zhang2018self} and ours. ADL excessively covers backgrounds because it simply encourages the model to use less discriminative parts, and SPG still over-estimates the bounding boxes although it tries to suppress background. In contrast, our method focuses on the entire object more accurately and estimates tighter bounding boxes. \fref{fig:figure4} illustrates more examples from our model. Our method not only spreads out of the most discriminative parts, but also restrains the activations in the object regions. Note that the water and mirrored image of the pelican does not earn large activation even though they are helpful cue for classification (the second row of the second column).

\section{Conclusion}
In this paper, we consider the background as an important clue for localizing the entire object without excessive coverage and present two novel objective functions. The crucial weakness of the previous methods is that they focus on discriminative parts rather than localizing the whole object, or extend too much on the background. The proposed contrastive attention loss guides the model to spread the attention map within the objects. The foreground consistency loss decreases the activation to backgrounds in the early layers. The generated attention map not only better localizes the target object but also suppresses the background concurrently. In addition, our non-local attention block enhances the attention map with a larger capacity to better optimize the proposed losses. We achieve state-of-the-art performance on ImageNet and CUB-200-2011 datasets and provide detailed analysis on the effects of our individual components.

\smallskip

\noindent \textbf{Acknowledgements.} This work was supported by the National Research Foundation of Korea grant funded by Korean government (No. NRF-2019R1A2C2003760) and Artificial Intelligence Graduate School Program (YONSEI UNIVERSITY) under Grant 2020-0-01361. We thank Junsuk choe for his valuable discussion.

\bibliographystyle{splncs}
\bibliography{egbib}

\end{document}